%% file: main.tex
\documentclass[acmtog]{acmart}

\setcopyright{acmlicensed}
\acmJournal{TOG}
\acmYear{2023} \acmVolume{42} \acmNumber{6} \acmArticle{269} \acmMonth{12} \acmPrice{15.00}\acmDOI{10.1145/3618354}

\usepackage{diagbox}
\usepackage{multirow}
\usepackage{mathtools}
\usepackage{bm}
\acmSubmissionID{419}
\usepackage{hyperref}

\usepackage{color}
\definecolor{turquoise}{rgb}{0.6,0.4,0}
\definecolor{purple}{rgb}{0.65,0,0.65}
\definecolor{dark_green}{rgb}{0, 0.5, 0}
\definecolor{green}{rgb}{0, 0.7, 0}
\definecolor{orange}{rgb}{0.7, 0.4, 0}
\definecolor{red}{rgb}{0.9, 0, 0}
\definecolor{brown}{rgb}{0.5, 0.16, 0.16}
\definecolor{blue}{rgb}{0,0,1}
\definecolor{teal}{rgb}{0.0, 0.4, 0.4}

\newcommand{\rz}[1]{{\color{black}\textbf{}#1}\normalfont}
\newcommand{\jz}[1]{{\color{black}\textbf{}#1}\normalfont}
\newcommand{\rh}[1]{{\color{black}\textbf{}#1}\normalfont}

\newcommand{\change}[1]{{\color{black}\textbf{}#1}\normalfont}

\begin{document}

%\title[TAP-Net++]{TAP-Net++: Learning }
\title[TAP-Net++]{Learning to Pack for Real: from Visual Sensing to Object Placement}
\title[TAP-Net++]{Neural Packing for Real: from Visual Sensing to Reinforcement Learning}
\title[TAP-Net++]{Complete Neural Packing: Visual Sensing to Reinforcement Learning}
\title[TAP-Net++]{Neural Packing: from Visual Sensing to Reinforcement Learning}

%% Authors
\author{Juzhan Xu}
\email{juzhan.xu@gmail.com}
\orcid{0000-0002-5132-238X}
\affiliation{%
  \institution{Shenzhen University}
  \country{China}
}

\author{Minglun Gong}
\email{gongml@gmail.com}
\orcid{0000-0001-5820-5381}
\affiliation{%
  \institution{University of Guelph}
  \country{Canada}
}

\author{Hao Zhang}
\email{haoz@cs.sfu.ca}
\orcid{0000-0003-1991-119X}
\affiliation{%
	\institution{Simon Fraser University}
	\country{Canada}
}

\author{Hui Huang}
\email{hhzhiyan@gmail.com}
\orcid{0000-0003-3212-0544}
\affiliation{%
	\institution{Shenzhen University}
	\country{China}
}

\author{Ruizhen Hu}
\email{ruizhen.hu@gmail.com}
\authornote{Corresponding author: Ruizhen Hu (ruizhen.hu@gmail.com)}
\orcid{0000-0002-6798-0336}
\affiliation{%
\institution{Shenzhen University}
\country{China}
}

\renewcommand{\shortauthors}{J. Xu, M. Gong, H. Zhang, H. Huang and R. Hu}

\input{figures/teaser}
\input{0-abs}

%%
%% Generate your CCSCML using http://dl.acm.org/ccs.cfm.
%%


%% Keywords

%% Teaser figure that appears on the top of the article.
%% Uncomment the includegraphics line to include an actual teaser image.
%% Make sure to fill out a description for accessibility
%\begin{teaserfigure}
%  \includegraphics[width=\textwidth]{images/teaser.png}
%  \caption{Teaser figure}
%  \Description{This is the teaser figure for the article.}
%  \label{fig:teaser}
%\end{teaserfigure}

\maketitle

%% import all math symbols
\input{math_utils}

%% The actual document with your content starts here

\input{1-intro}

\input{2-related}
\input{3-method}
\input{4-exp}

\input{5-conclusion}

%% Acknowledgements
\begin{acks}
\jz{We thank the anonymous reviewers for their valuable comments. Thanks are also extended to Jiahui Zhu and Xueqi Ma for their help. This work was supported in part by NSFC (62322207, U2001206, 62161146005, U21B2023), Guangdong Natural Science Foundation (2021B1515020085), Shenzhen Science and Technology Program (RCYX20210609103121030), DEGP Innovation Team (2022KCXTD025), NSERC (No.~611370, 2023-05035), gift funds from Adobe, and Guangdong Laboratory of Artificial Intelligence and Digital Economy (SZ).
}
\end{acks}
% copy from tapnet: NSERC Canada (611370, 293127),  
%% Bibliography.
%% Uncomment the bibliography line and link to an actual bib file
\bibliographystyle{ACM-Reference-Format}
\bibliography{bibliography.bib}

%\clearpage

%% Appendix
%\appendix

%\input{6-appendix}
%\section{Appendix}
%NOTE: appendices should NOT be included in the submission, and instead, should all go to the supplementary in the submission.
\end{document}

%% file: figures/teaser.tex
\begin{teaserfigure}
	\centering
	\includegraphics[width=\linewidth]{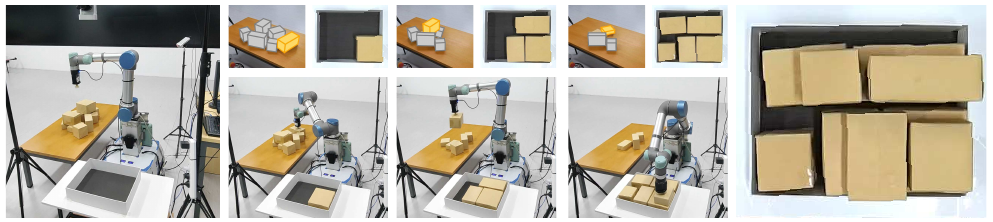}
	\caption{We \rz{present a full solution pipeline for the 3D transport-and-packing (TAP) problem, which is based on reinforcement learning and} real-time RGBD sensing of the source objects and the packing state in the target container (top row). The series of photographs show a Universal Robots UR5e robotic arm executing a TAP solution produced by our network, which simultaneously selects an object (colored in yellow) to pack and determines a final packing location. %\mlc{do we have results for boxes of real products and of different dimensions, such as iPhone, iPad, VR lens.} \rh{[No, Juzhan is not at SZU right now and we don't have new videos for the real test. The new video we have from the real working environment has a different problem setting as in Figure~\ref{fig:real_env} ]}
	}
	\label{fig:teaser}
\end{teaserfigure}

%% file: 0-abs.tex
\begin{abstract}
We present a novel learning framework to solve the %\rh{complete}
%real-world 
transport-and-packing (TAP) problem in 3D.
It constitutes a full solution pipeline from partial observations of input objects via RGBD sensing and
recognition to final box placement, via robotic motion planning, to arrive at a compact packing in a 
target container. 
The technical core of our method is a neural network for TAP, trained via
{\em reinforcement learning\/} (RL), to solve the NP-hard combinatorial optimization problem. Our
network simultaneously selects an object to pack and determines the final packing 
location, based on a judicious encoding of the continuously evolving states of partially observed source 
objects and available spaces in the target container, \change{using separate encoders both enabled with 
attention mechanisms.}
\change{The encoded feature vectors are employed to compute the matching scores and feasibility masks of 
different pairings of box selection and available space configuration for packing strategy optimization.}
Extensive experiments, including ablation studies and physical packing execution by a real robot (Universal Robot UR5e), 
are conducted to evaluate our method in terms of its design choices, \change{scalability}, generalizability, %tion capabilities,
and comparisons to baselines, including the most recent RL-based TAP solution.
\change{We also contribute the first benchmark for TAP which covers a variety of input settings and difficulty levels.}
\end{abstract}

%% file: math_utils.tex
% define all math-related symbols as global commands here

\newcommand{\exemplar}{ E }

%% file: 1-intro.tex
\section{Introduction}
\label{sec:intro}

3D packing is a real-world problem in the transport and warehousing industries, with
% With the surge of e-commerce sales especially since the start of the pandemic, 
ever-increasing number of packages delivered daily.  According to a study by the National Institute of Standards and Technology (NIST) 
in the US, packaging represents about 25\% of the solid waste generated annually. 
%A study by the Ellen MacArthur Foundation also estimated that by 2050, plastic packaging alone could account for 20\% of the global oil consumption and 15\% of the global carbon budget.  
Hence, improving packing efficiency has important economical and environmental impacts.
On top of that, there has been a continuing drive towards robotic automation of warehouse operations including
object transport and packing for improved operational continuity and worker safety, e.g., when
handling heavy packages or packing in a confined space.
%\change{Figure~\ref{fig:real_env} shows a real-world workspace populated by boxed products to be packed, where such an automation 
%	would be highly desirable.}
%\jzc{Should we remove the figure 2 to simplify the introduction?}

% frequently encountered in several industry settings including shipping and delivery services for transportation and eCommerce, as well as assembly line operations.
%Every year, significant amount of solid waste results from inefficient packing solutions which led to  unnecessary high packaging costs.
%Every year, at least 100 million tons of solid waste globally comes from packaging waste, in which partly created by inefficient packaging. 

%\paragraph{Packing vs.~transport-and-packing}
%
To date, most packing problems tackled in graphics, whether for chartification~\cite{levy2002least,noll11,sander2003multi,limper18_boxcutter},
%aesthetic 
pattern generation~\cite{kaplan2000,doyle2019}, % nagata2020
or fabrication~\cite{koo2016towards,vanek2014packmerger,chen15_dapper},
have only focused on maximizing the efficiency of the {\em final packing state\/}. 
The recent work by Hu et al.~\shortcite{hu2020tap}, TAP-Net for Transport-and-Pack, was the first
to jointly account for the {\em transport\/} aspect of the problem, i.e., how to select an object to 
pack and how to transport it to the target container, as both issues must be considered for physical 
packing. %, e.g., for package delivery at a postal or eCommerce branch, since the initial physical configuration
%of the packages is directly connected to the packing problem itself.
%
%To our knowledge, TAP-Net was the first neural network designed to tackle the transport-and-pack problem.

As a first solution however, % at transport-and-pack, an NP-hard combinational optimization problem, 
TAP-Net makes simplifying assumptions on the number, %size,
pose, and location of the source objects and resorts to heuristic
packing strategies, rendering the problem setting and solution impractical and sub-optimal.
Our goal in this work is to develop a learning framework which encompasses a \rz{{\em full TAP pipeline\/}},
%real-world packing
from partial observations of input objects via visual sensing to final box 
placement to arrive at a compact packing,
%While TAP-Net assumes that the sizes and locations of all source objects are known ahead of time and the target container is initially empty, 
without making limiting assumptions on the object or container states as in TAP-Net, %~\cite{hu2020tap},
except that all objects \rz{remain} as {\em boxes\/}; see Figure~\ref{fig:teaser}. 

Figure~\ref{fig:overview} shows an overview of our pipeline. 
%
%In our current solution, off-the-shelf methods are 
%employed for box recognition from RGBD images %, namely, Mask-RCNN~\cite{he2017mask}, 
%and motion planing for the robot arm to realize the predicted packing. %~\cite{gammell2020batch}. 
%The technical core is referred to as TAP-Net++, which is trained to select boxes to pack as well as to 
%determine the packing positions in the target container.
%
%\section{Overview}
%\label{sec:overview}
%
Given a set of source objects stacked on a workspace, our goal is to pack the objects compactly into a 
target container using a robot arm with a vacuum gripper. %We refer this setup as the workspace for the robot arm.
Two RGBD cameras are employed to observe source objects and the target container, respectively, for object 
selection and packing strategy optimization, as shown in Figure~\ref{fig:overview}(a).
	 %detect the workspace, one camera takes a single-view RGBD image of source objects to detect the observed boxes, and another one is used to observe the state of the target container.
	
	%We set a robot arm with a vacuum gripper for objects' pick and place, a set of source objects stacked in front of the robot, a container is placed on the left as the target. Our goal is to pack the object compactly with the robot into the target container. The objects we handle are abstracted by axis-aligned boxes (AABBs), possibly rotated by $90^{\circ}$, into the target container, which is also a box itself. We allow the robot to pick the objects from various grasp direction, and finally packing the object into the target container from the top.
	
With the captured RGBD images, we first analyze the current workspace and extract information important for packing, as shown in Figure~\ref{fig:overview}(b). For source objects that are observable from the camera, we estimate their dimensions and further extract a {\em precedence graph\/}, which encodes the constraints on the pick-up orders among the objects. For the target container, we extract its {\em height map\/}, which is used to generate a set of {\em empty maximal spaces\/}, or EMS~\cite{parreno2008maximal} for short, to indicate candidate packing locations. %More details about the workspace analysis can be found in Section~\ref{sec:analysis}. %gonccalves2013biased

The technical core of our method is the packing network, coined TAP-Net++. It takes both box information and the set of candidate EMS as input, 
%and selects the optimal pair of box state and candidate EMS, i.e., one with the highest matching score, to guide the packing, as shown in Figure~\ref{fig:overview}(c). 
\change{and simultaneously selects a box to pack and determines the final packing 
location, based on a judicious encoding of the continuously evolving states of partially observed source 
boxes and EMS configurations, using separate encoders both enabled with 
attention mechanisms (see Figure~\ref{fig:network}). The encoded feature vectors are employed to compute the matching scores and feasibility masks of 
different pairings of box state and candidate EMS. Finally, the pair with the highest matching score is selected to guide the packing,
as shown in Figure~\ref{fig:overview}(c).}
%
%Note that some of the invalid packing pairs are masked out using dashed lines, and the one with the highest matching score in the remaining pairs is selected. 
%To handle dynamic workspace information, TAP-Net++ has to be able to handle different numbers of input, and thus we adopt a transformer-based encoder. 
With the packing problem an NP-hard combinatorial optimization, which is unfitting for supervised learning due to the difficulty of collecting sufficient ground-truth data, we train TAP-Net++ using reinforcement learning (RL). 
Importantly, TAP-Net++ is trained for both box selection and packing into the target container, while TAP-Net only learns the former and determines the packing
position separately and heuristically.

%\input{figures/real_env}

%	More details about the network design and training can be found in Section~\ref{sec:network}.
	
Once a box has been selected and the target packing position determined by TAP-Net++, we employ a motion planning method %~\cite{gammell2020batch} 
to generate the motion sequence for the robot arm to execute the packing. As shown in Figure~\ref{fig:overview}(d), the workspace is updated accordingly to initiate the next box selection and packing.

Extensive experiments \rh{including ablation studies} are conducted to evaluate our method in terms of its design choices, \change{scalability, and generalizability. %generalization capabilities,
%and comparisons to baselines. %, including the closest competitor TAP-Net. %recent RL-based TAP solution.
%
We also contribute the first benchmark for TAP which covers a variety of input settings and difficulty levels, over which we comprehensively
evaluate TAP-Net++ against TAP-Net, its closest competitor.
\rh{Physical packing execution by a real robot, a Universal Robot UR5e, has also been carried out to demonstrate deployment of our method in a non-simulation setting.}
}

	%Based on the images from the cameras, we analyze the configuration, estimate the observed boxes and their precedence constraints, as well as the height map of the target container, we can further compute the empty maximal spaces (EMS)\cite{gonccalves2013biased} of the container as candidate packing positions, this information would be the input of our network module. Our network tries to find the best box state and its corresponding packing position by computing the probability of each packing strategy pair. The box state and position pair with the highest score would be chosen as the strategy of the current step. 
	
	%We apply the existing planning algorithm controlling the robot to pick the goal box. To ensure the size of the selected observed box is correct, we use a camera to take an RGBD image of the selected box for second-time detection. If the size of the observed box is correct, we can place the box at the expected packing position to update the workspace. Otherwise, we apply a simple heuristic packing algorithm~\cite{karabulut2004hybrid} to compute the final packing position with the real box size. This process repeats until all the source boxes are packed into the target container.

\input{figures/overview}

%% file: figures/overview.tex
\begin{figure*}[!t]
    \centering
    \includegraphics[width=0.99\textwidth]{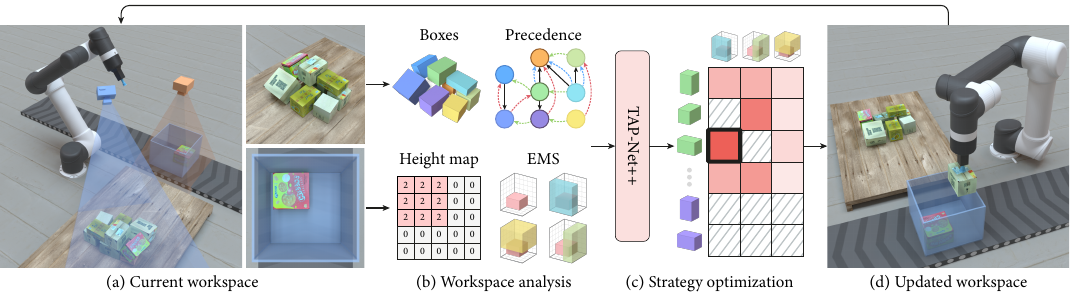}
\caption{Overview of our method. Given the RGBD images captured from both the source objects and the target container (a), our method starts with extracting essential information for packing optimization. This includes the dimensions of source objects and their precedence constraints, as well as the height map of the target container and candidate packing locations, referred to as EMS (b). This information is fed into TAP-Net++, which outputs a matching score for each pair of object state and candidate EMS. The pair with the highest matching score, highlighted with a thick border, is selected as the optimal packing strategy (c), i.e., the third box state will be selected and packed into the first EMS. The robot arm then picks the selected box with the desired state and packs it into the optimized position, which triggers the next iteration of object selection and packing. }
\label{fig:overview}
\end{figure*}

%% file: 2-related.tex
\section{Related work}
\label{sec:related}
 
In this section, we \rz{briefly mention several real-world examples of automated packing using robots to motivate the TAP problem}.
We then cover prior packing strategies and provide a detailed comparison between TAP-Net++ and TAP-Net~\cite{hu2020tap}. %, the most closely related prior work.

\begin{table}[t]
\caption{Comparing various aspects of TAP-Net++ vs.~TAP-Net.}
\label{tab:comp_TAPNet}
\begin{center}
\resizebox{1.0\linewidth}{!}{
\begin{tabular}{lcc}
% \toprule
 & TAP-Net & TAP-Net++ \\
\midrule
\multirow{3}{*}{Input} & Virtual boxes, & Real boxes,  \\
 & axis-aligned, and & casually stacked, \\
 & packed in a container & with random poses \\
\midrule
Box config. & Known & Predicted from partial observation  \\
\midrule
\# instances & Fixed & Flexible  \\
\midrule
Overall solution & Two-step optimization & One-step optimization  \\
\midrule
Network output & Packing order and state&  Packing order, state, {\em and location\/} \\
\midrule
Packing strategy & Heuristic: MACS & Optimized by neural network \\
\bottomrule
\end{tabular}
}
\end{center}
\end{table}

\vspace{-5pt}

\paragraph{Real-world robotic automation.}
\rh{Automating warehouse operations including order retrieval, shelfing, packaging, and packing is one of
the most critical innovations considered by many commercial companies including eCommerce giants 
such as Amazon and Walmart. 
%
%The Alphabot systems developed by Alert Innovation has been deployed by
%Walmart to automate and speed up order picking. The system takes customer online orders and relies on
%autonomous carts to retrieve the ordered items and delivers them to human-operated workstations to 
%check, package, and deliver the final order. 
%
At many of its fulfillment centers, Amazon employs its warehouse robots to automate picking, moving, and 
packing solutions. In particular, two of Amazon's recently unveiled robotic arm cells, Cardinal and Sparrow, 
are designed for the transport-and-pack problem. While the main innovation of the Sparrow robot is to use 
its suction cups to firmly grasp and move items of varying shapes and sizes, the task carried out by the 
Cardinal robotic arm is the closest to ours. According to this 
\href{https://www.youtube.com/watch?v=3dB-YHAP5RM&t=48s}{\color{black}{\underline{official Youtube video}}} from Amazon,
Cardinal employs ``advanced artificial intelligence and computer vision to nimbly and quickly select one 
package out of a pile of packages, lift it, read the label, and precisely place it in a GoCart.'' However,
it does not appear that the box selection or packing was optimized to arrive at a compact packing
at the target containers. }

\vspace{-5pt}

\paragraph{Packing strategies}
One strategy adopted by prior works is to traverse all possible packing positions in the container and
pick the one with the highest score, where the scoring criterion may be Deepest-Bottom-Left with Fill 
(DBLF)~\cite{karabulut2004hybrid} or via height map minimization~\cite{hu17,wang19_1}.
%\rhc{It seems that those two are mainly used to guide the traverse of all possible packing positions instead of determining the scoring criterion.}
%
Another way is to select from a set of candidate packing positions. To characterize such positions,
corner points~\cite{martello2000three, lodi2002heuristic, martello2007algorithm}, extreme points~\cite{crainic2008extreme, baldi2012generalized}, and empty 
maximal spaces (EMS)~\cite{parreno2008maximal} have been employed. In particular, heuristic  %parreno2008maximal
methods such as Best-Match-First (BMF)~\cite{li2015hybrid} and Online Packing Heuristic (OnlineBPH)~\cite{ha2017online} have proposed 
suitable rules to select packing locations based on EMS.
In our work, we extract a heigh map over the target container and generate candidate packing locations via EMS.
The discretized description of the container states significantly reduces the searching space for packing.

Among the recently proposed learning-based packing strategies, Zhao et al.~\shortcite{zhao2021online} designed a
neural network to score all possible packing positions based on height maps over the container, while in \cite{zhao2021learning},
a hierarchical packing configuration tree is learned based on EMS.
Hu et al.~\shortcite{hu2020tap} trains a neural network, TAP-Net, to optimize the order of boxes to be packed, while resorting
to a heuristic coined Maximize-Accessible-Convex-Space (MACS) as the packing strategy. MACS was designed to account for
accessibility issues pertaining to future packing and to accommodate large objects. They opted to not train a
separate packing network as it did not improve over their heuristic-based approach.

% RZ: not sure I really understand the point here or why this is stated here.
%
%Note that although height maps have been widely employed to represent container states for packing analysis, they are only discrete approximations of the container state in continuous space, and the searching space size is closely related to the quantization level of the container.

%Since the set of object states is discretization, to optimize packing states and positions at the same time, we prefer the packing positions also be discretization, so that we can easily match the object states and packing positions. In several concepts about packing position, EMS and height map can be transformed into each other, and it describes the state of the container in a discretized way, which significantly reduces the searching space of packing position. So we adopt EMS as the representation of containers in our work.

%e.g., 
%Deepest-Bottom-Left with Fill (DBLF)~\cite{karabulut2004hybrid}, height map minimization~\cite{hu17,wang19_1},
%as well as Empty-Maximal-Space (EMS)~\cite{ramos16} which we have adopted in our work. 
%\rz{[RZ: I would discuss some motivations for this choice and add other relevant discussions on packing strategies.]}

\vspace{-5pt}

\paragraph{TAP-Net++ vs.~TAP-Net}
Table~\ref{tab:comp_TAPNet} compares TAP-Net++ and TAP-Net in terms of input, 
output, box configuration, and packing strategies, etc.
In more detail,
%First of all, the input to TAP-Net is a set of axis-aligned virtual boxes packed in a container with the assumption that all the box configurations are known, while TAP-Net++ considers more realistic cases where real boxes are casually stacked with random poses and only partial observations can be captured for box detection and pose estimation.
%Moreover, 
TAP-Net is designed to only deal with a fixed number of instances as input, and use an auxiliary rolling scheme to handle tasks with more instances. With a more advanced network architecture, TAP-Net++ affords more flexibility to the number of instances and is able to deal with dynamic inputs obtained from partial observations.
More importantly, TAP-Net adopts a two-step optimization, where the network only optimizes the packing order and state of the input boxes and the final packing position of the selected box is determined by a heuristic packing strategy, which may lead to suboptimal solutions. TAP-Net++ discretizes the packing space and directly outputs the packing location, along with the parking order, through a one-step optimization.

%\subsection{Pick and Place task}
%\subsection{Pick and Pack task}
%\subsection{Neural Optimization}

%% file: 3-method.tex
\section{Method}
\label{sec:method}

Here we discuss the two key steps of our method in more detail: the analysis of workspace state in Section~\ref{sec:analysis} and the optimization of packing strategy in Section~\ref{sec:network}.

\subsection{Workspace state analysis}
\label{sec:analysis}

\paragraph{Source objects and precedence graph}
The objective of this phase is to identify objects that need to be packed and extract their precedence relations using the single-view RGBD image captured; see the blue camera in Figure~\ref{fig:overview}(a).
Object detection is performed using the Mask RCNN network~\cite{he2017mask}, which is followed by the estimation of dimensions and 6D pose for each object using the cuboid fitting algorithm~\cite{jiang2013linear}.
We allow the robot arm to grasp each box object along its three axes, and possibly rotate it by $90^{\circ}$ before packing into the target container. Consequently, there are six distinct packing states for each object; see Figure~\ref{fig:preced}(a). %For a total of $n$ observed boxes, we need to choose among $6n$ object packing states, same as the setting for 3D in~\cite{hu2020tap}. 
	
For a given box, certain packing states may not be accessible to the robot arm due to obstructions caused by other boxes. Thus, it becomes essential to extract a precedence graph that captures the relationships between the boxes. Inspired by the approach proposed in~\cite{hu2020tap}, we assign each box as a graph node, and directed edges represent different precedence relations among the boxes. However, unlike the assumption made in~\cite{hu2020tap} where all source objects are assumed to be axis-aligned and initially packed in a container, we allow the objects to be casually stacked in the workspace with random poses that may not be axis-aligned. Hence, the extraction of precedence relationships becomes more complex.

	More specifically, we define two types of precedence: one is Movement Block (MB), indicating the movement of an object $O_1$ is blocked by another object $O_2$ since $O_2$ is at least partly on top of $O_1$; the other is Access Block (AB), which indicates one of object $O_1$'s surfaces cannot be reached by the robot arm due to another blocking object.  Based on the direction of the surface that are blocked, AB can be further divided into XAB, YAB, and ZAB, where X, Y, Z axes are randomly determined for each box during the cuboid fitting step.  Since a box can be picked up from either side for a given direction, we store the blocking side of each AB edge. Then $O_1$ can be picked up from its $X$ (or $Y$/$Z$) axis direction as long as there is no MB edge and at least one side of $O_1$  that is perpendicular to $X$ (or $Y$/$Z$) axis direction has no XAB (or YAB/ZAB)  edges pointing to it.

	Figure~\ref{fig:preced} shows the process of precedence extraction on an example scene with three stacked boxes. 
	%We show the six different packing state of the yellow box and the grasping along three different axis directions to enable those packing states in Figure~\ref{fig:preced}(a).
	To determine Movement Blocks, we project all the boxes from the top and check whether the projections of each pair of objects are overlapped as in Figure~\ref{fig:preced}(b). If so, an MB (black) edge from the object on top to the one on bottom is added; see the three MB edges in Figure~\ref{fig:preced}(c).
	For Access Blocks, we check whether the robot arm has access to the two surfaces along each axis direction. As shown in (a), the blue gripper along the local z-axis of the yellow block is blocked by the orange box. Hence a ZAB (blue dashed) edge is added from the orange to the yellow nodes; see Figure~\ref{fig:preced}(c). 
	Moreover, since the bottom of the yellow box is inaccessible, a ZAB edge from the yellow node to itself is also added. 
	Other precedences can be derived in a similar way and the full precedence graph of this example is presented in (c).
	
	\input{figures/precedence}

\change{Suppose we have detected $n$ objects denoted as $\{O_i\}_{i=0}^{n-1}$ in the current workspace. The extracted precedence information for each object $O_i$ can be represented as $P_i =[P_i^\text{M}, P_i^\text{XA}, P_i^\text{YA}, P_i^\text{ZA} ]$, where $P_i^\text{M} \in \{0,1\}^n $ indicates whether there is an MB edge pointing from each of the $n$ objects to  $O_i$ and $P_i^\text{\{X/Y/Z\}A}  \in \{0,1\}^n $ indicates \{X/Y/Z\}AB edge pointing to  $O_i$ along one of $O_i$'s local axes.
	
During the packing process, we need to simultaneously decide which object to pack next and its orientation in the target container, i.e., the object's packing state. Hence, we extend the precedence information to the packing state level, represented as $\{p_{6i+s}\}$, where $s\in[0,5]$ corresponds to each of the six distinct packing states shown in Figure~\ref{fig:preced}(a). Depending on the packing state $s$, the accessibility of object $O_i$ depends on $P_i^\text{M}$ and only one of $\{P_i^\text{XA}, P_i^\text{YA}, P_i^\text{ZA}\}$. This is because whether $O_i$ can be picked up and packed along one of its axes is not affected by access blocks along two other axes. Therefore, we set packing state precedence $\{p_{6i+s}\}=[P_i^\text{M}, P_i^\text{sA}]\in \{0,1\}^{2 \times n}$, where $P_i^\text{sA}$ is the AB precedence corresponding to the direction of packing state $s$. 

Finally, to fully encode all information associated with object $O_i$'s packing state $s$, a 3D vector $b_{6i+s} \in \mathbb{R}^3$ is used to store the dimension of $O_i$ under the orientation specified by state $s$. Basically, $b_{6i+s}$ indicates the minimum empty space needed in the target container to pack $O_i$ along orientation $s$. $b_{6i+s}$ and  $p_{6i+s}$ together form the inputs to the source (object) encoder, i.e., $B_{6i+s} = [b_{6i+s}, p_{6i+s}]$.
To simplify the representation on the following packing strategy discussion, we reindex all the object states as $\{B_j\}_{j=1}^N$, where $B_j = [b_j, p_j]$ and $N = 6n$ represents the total number of object states.}

\input{figures/ems}

\input{figures/network}

\paragraph{Target container and EMS}
	Our goal here is to generate a set of candidate locations for packing based on the current container state. This is achieved through computing candidate empty maximal spaces (EMS). EMS provides a simple yet effective placement rule, which is to align the deepest-bottom-left (DBL) corners of the selected object box and the EMS.
	
	As shown in Figure~\ref{fig:overview}(a), the camera in orange provides a single-view RGBD image of the target container. We first project the RGBD image to the local coordinates of target container to form a 2D height image, which is then discretized into a grid map to simplify candidate EMS calculation. 
	The height of each grid is set to the maximum height of the corresponding area in the height image, resulting the grid map being axis-aligned with the local coordinate of the target container.

\change{An EMS is defined as the largest empty orthogonal space whose size cannot extend further along the three axes \cite{parreno2008maximal}.
	In previous approaches, EMS are only generated from a fixed corner of the container. To optimize packing, our approach considers more candidate locations through generating EMS from the left-bottom corner of each visible object; see Figure~\ref{fig:ems}.
%	Note that for a target container with varying heights, we can generate multiple EMS. 
		
	Suppose a set of candidate EMS $\{E_k\}_{k=1}^{M} $ is constructed, then each EMS $E_k$ can be represented by its DBL corner $[c_k^x, c_k^y, c_k^z]$ and its dimension $[d_k^x, d_k^y, d_k^z]$. That is, $E_k = [c_k^x, c_k^y, c_k^z, d_k^x, d_k^y, d_k^z]  \in \mathbb{R}^6 $.
	}

\subsection{Packing strategy optimization}
\label{sec:network}

Different from~\cite{hu2020tap}, where the network only selects the object/orientation to pack and then uses heuristics to find the optimal packing position in two separate steps, we design a network to simultaneously select and pack the object in one step. 
This is done by finding the $\langle B_{j}, E_k\rangle$ pair with the highest matching score under a newly defined feasibility constraint.

%\paragraph{Network architecture}
	%After the analysis of workspace, we get a set of box states and a set of EMSs, we define each pair of a box state and a EMS as a transport-and-pack strategy. The goal of our network is to find an optimal strategy at each step.
	
Figure~\ref{fig:network} shows the structure of our network. It can be divided into two parts: a source encoder encodes the box sizes and their precedence information into high-dimension features, and a target encoder learns the embedding of each candidate EMS. \change{Then we multiply the features from both source and target encoders to compute the matching score and the feasibility of each pair of $\langle  B_{j}, E_k \rangle$. The pair with the highest multiplied score will be selected as the output of the network. }
The box state will guide the robot to pick the selected box from the selected side and pack it at the DBL corner of the corresponding EMS in the target container.
Followings, we give more details about each key component of our network and explain how the network is trained.

\paragraph{Source encoder}
\change{As shown in the source encoder module in Figure~\ref{fig:network}, taking the set of object packing states  $\{B_{j}\}_{j=1}^{N}$ as input, we first extract the corresponding embedded features separately using the object encoder, and then further use an attention mechanism to extract core information in the feature map. Since the box state is order dependent in the precedence matrix, we apply a position encoding on each embedded feature before passing it to the attention mechanism. The attention module will then extract the final source object features $\{S_{j}\}_{j=1}^{N}$ with $S_{j} \in \mathbb{R}^{D}$ as output.
More details about the object encoder can be found in the supplementary material.

\paragraph{Target encoder}
	%Each EMS can be represented by its DBL corner and size $e_j = (p_j^x, p_j^y, p_j^z, s_j^x, s_j^y, s_j^z)$,  where the whole set of candidate EMS encodes the free space information of target container. 
	For a set of $M$ candidate EMS $\{E_k\}_{k=1}^{M}$ extracted in the current target container, we first use an MLP layer as the space encoder to extract features, and then use an attention module  %as in the source encoder 
	to get the final target EMS features $\{T_k\}_{k=1}^M$ with $T_k \in \mathbb{R}^{D}$.
	%\mlc{the number of EMS can be different at different steps.  How to you handle the variation?  Do you assume a max number of EMS?} \jzc{The attention module allows the network to handle a dynamic number of EMS, we assume a max number of EMS as $5N_{obj}$ during training, the data of the rest of EMS are all padded with zeros. We use an EMS mask in attention module to prevent the feature computation between valid EMS and padding zeros.} 

\paragraph{Packing pair selection}
Given source object features $\{S_j\}_{j=1}^N$ and target EMS features $\{T_k\}_{k=1}^M$, we compute a selection probability map $\mathcal{P} \in [0,1]^{N \times M}$ to select the $\langle B_j, E_k \rangle$ pair with the highest probability for packing strategy optimization.
In more detail, we first compute the matching score matrix $\mathcal{M}  \in  \mathbb{R}^{N \times M}$ with $\mathcal{M}_{jk} =  \langle S_j, T_k\rangle$, where $\langle\cdot,\cdot\rangle$ represents an inner-product.
We then pass both $S_j$ and $T_k$ through an MLP layer and get their inner product to construct the feasibility mask $\mathcal{F} = \mathbf{sigmoid}( \mathcal{F}_{jk})$, where $\mathcal{F}_{jk} = \langle \mathbf{MLP} (S_{j}), \mathbf{MLP} (T_k)\rangle $. 
Note that since a box state can not be packed into a smaller EMS or may not be accessible due to precedence constraints, we further compute a corresponding validity mask $\mathcal{V}\in\{0,1\}^{N\times M}$ to filter out invalid pairs. 
The final selection probability map is then defined as $\mathcal{P} = \mathbf{softmax}(\mathcal{M}  \odot \mathcal{F}  \odot \mathcal{V})$, where $\odot$ represents the element-wise product.}
% and $A_{jk}$ represents the matching score between box state $B_j$ and EMS $E_k$. 

%to optimize the packing strategy, we predict a matching score matrix $A$ to estimate  
%and a feasibility mask $F$}
%the strategy matrix $A\in R^{N\times M}$ is constructed with $A_{jk}=\langle S_j,T_k\rangle$, where $\langle\cdot,\cdot\rangle$ represents an inner-product and $A_{jk}$ represents the matching score between box state $B_j$ and EMS $E_k$. 

%The final matching matrix would be $A^{'}=A \odot V$, where $\odot$ represents the element-wise product. The entry with the highest value in  $A^{'}$ corresponds to the $\langle$oriented box, EMS$\rangle$ pair for the robot arm to execute.

Note that in some cases, a box to be picked up may have its sides occluded by other boxes from the viewpoint of the source camera. Hence the estimated dimensions may not be accurate, which could lead to improper packing decisions. To address this issue, we always reevaluate the dimensions of an object after it is picked up by the robot but before it is packed into the target container. If the initially estimated dimensions are accurate, the object will be placed in the container as planned. Otherwise, we re-run the TAP-Net++ using the updated dimensions and select the EMS entry with the highest score under the current box state constraint.  This allows the robot to pack the already picked-up box into an optimal EMS location, without the complication of putting down the box and grabbing another box or another surface of the same box. %\mlc{I think this will work.  Pls verify.}\jzc{I have updated the results, the performance is similar to the previous method using a heuristic packing strategy.}
\change{This process repeats until no more source objects can be packed into the target container or the last packed object falls off due to unstable placement. In the latter scenario, the packing process for the current container is terminated prematurely, effectively penalizing the placement of objects in unstable locations.}

\paragraph{Network training} 
\change{We use the PPO algorithm~\cite{schulman2017proximal} to train our network due to the lack of data for supervision. 
To measure the final packing quality, we use compactness $C$ as the reward function to guide the training.
The compactness $C$ is defined as the ratio between the total volume of packed boxes and the container volume. 
Intuitively, the compactness measure favors tightly-packed boxes and $C=1$  if and only if the container is fully filled. % by boxes up to a flat top surface.
If there are multiple target containers are used, we use the mean compactness of all the containers $C= {\sum_{i=1}^{N_t}{C_i}} / N_t$ as the reward function, where $N_{t}$ is the number of containers used.}

%% file: figures/precedence.tex
\begin{figure}[!t]
    \centering
    \includegraphics[width=0.49\textwidth]{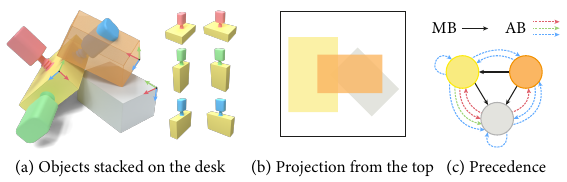}
\caption{Precedence extraction. (a) Example scene with three objects stacked on the desk, where the assigned X, Y, Z directions are shown in red, green, blue arrows. Grippers picking the yellow box along three directions are shown on the left, whereas all six possible packing states are shown on the right; (b) Projection of the scene from the top for MB edge detection;  (c) The final precedence graph of the example scene, where MB edges are shown in black solid arrows and AB edges for different directions are shown in dashed arrow with different colors.}
%\mlc{Should the dashed edge from orange to grey be red color?}}
\label{fig:preced}
\end{figure}

%% file: figures/ems.tex
\begin{figure}[!t]
    \centering
    \includegraphics[width=0.49\textwidth]{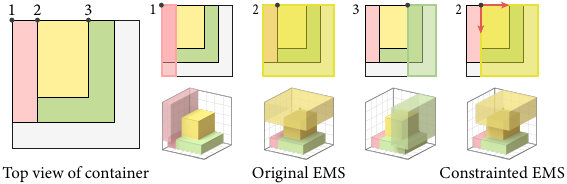}
	\caption{\change{Candidate EMS generation. The map on the left shows the top view of the container, with three \rh{representative} top-left corners found on the map highlighted.
	% \jz{(Only a representative subset of candidate EMS is shown here)}. %\mlc{corners are not highlighted in the height map}.
	Existing approaches compute the original EMS by expanding along all directions as shown in the middle three examples.
	Our approach extracts more candidate packing locations by constraining the expanding directions indicated by the red arrows,
	% \mlc{more candidate location and constraint on expanding directions may not correlated.  Also, are there results show that constrained EMS works better?}, 
	which results in a different EMS for the same corner \#2.
	Note that due to the blocking of the tall yellow box, the constrained EMS for corner \#1 and \#3 is exactly the same as the original EMS.}
}
\label{fig:ems}
\end{figure}

%% file: figures/network.tex
\begin{figure*}[!t]
    \centering
    \includegraphics[width=0.99\textwidth]{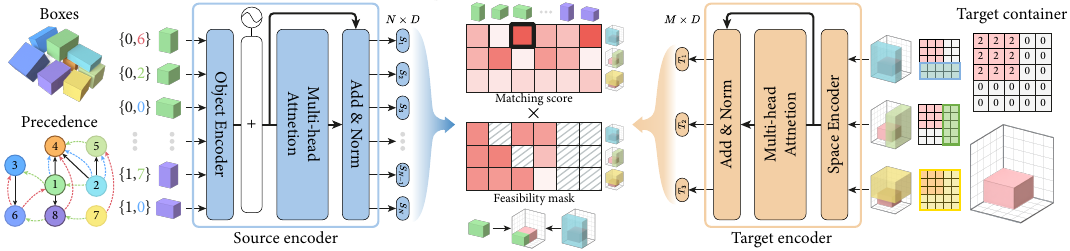}
\caption{Network structure of TAP-Net++. A source encoder is used to encode different box states and their precedence relations, whereas a target encoder encodes candidate EMS in the container. \change{The feature vectors are further used to compute the matching score and feasibility mask of different $\langle$oriented box, EMS$\rangle$ pairs for packing strategy optimization.}}
\label{fig:network}
\end{figure*}

%% file: 4-exp.tex
\section{Experiments}

\subsection{TAP-Benchmark}

\rz{To thoroughly study the TAP problem and evaluate its solutions, we establish a benchmark consisting of multiple source box settings and target container settings, which have been motivated by 
different %real-world 
applications and usage scenarios; see Figure~\ref{fig:scene_setting}.}

\change{
In more detail, there are three different settings for the source boxes with different levels of flexibility and difficulty: \\
$\bullet$ \textbf{FIX}: all source objects are sampled from a fixed set of $N_f$ objects.\\ 
$\bullet$ \textbf{RAND}: the source object sizes have more variations and are randomly sampled from a given distribution. \\
$\bullet$ \textbf{PPSG}: short for Perfect Packing Solution Guaranteed, refers to a set of random boxes that are guaranteed to have a packing solution that can fully pack all the objects in one container, resulting in a perfect packing with $C=1$. The data construction method is based on the bin packing problem generator~\cite{laterre2018ranked}.%The details about how to construct the PPSG sets can be found in the supplementary material. 

There are two different settings for the target containers with different packing goals and constraints: \\
$\bullet$ \textbf{Single}: there is only a single target container and the goal is to fill the container with the source objects as fully as possible.\\ 
$\bullet$ \textbf{Multi}: a new empty target container can be added once the current container either runs out of space or is filled with unstable packing, and the goal is to pack all the source objects into the target containers. When there are multiple containers, we further subdivided the setting into two cases: \textbf{Multi-All} where all containers can be used to pack new objects, and \textbf{Multi-Last} where only the last added container can be used for future packing. 
%$\bullet$ \textbf{Multi-Last}: \\
%During packing tests, once there is not enough space for the currently selected box to pack, a new empty target container will be added for packing. 
%We use the mean compactness of all the containers $C={\sum_{i=1}^{N_{\text{ctn}}}C_i} / {N_{\text{ctn}}}$ as the evaluation metric of packing quality,  where $N_{ctn}$ is the number of containers used.
%\jz{ Figure~\ref{fig:scene_setting} shows the example scene of different packing goals and constraints.} \rh{[Do we need to find figure for different input settings or it's easy to explain using text?]}

We set the target container size as  $100 \times 100 \times 100$, and generate $N_{s}$  source objects for packing, with the dimensions of each object being multiple of a unit length $u = 1$ and smaller than $100$.
For the \emph{Single} container setting, we use the compactness $C$ as the evaluation metric.
For the \emph{Multi} container setting, as multiple containers are used, we also \rz{add} the container number $N_t$ as an evaluation metric, \rz{along with compactness}. 
Moreover, we further define the ideal container number for the set of source objects $\{O_i\}_{i=1}^{N_s}$ as $N_t^* = \lceil \sum_i{V_i} / V_t \rceil$, where $V_i$ and $V_t$ are the volumes of the object $O_i$ and the target container, respectively, and use the extra number of containers used $\Delta N_t = N_t - N_t^*$ as an evaluation metric.

In the following experiments, we show results under different combinations of input-output settings as well as comparisons to baselines to show the robustness and generality of our method.}

\input{figures/scene_setting}

\subsection{Baseline algorithms and comparisons}

\paragraph{Baseline algorithms} 
We here compare our method to two baselines for object selection and packing optimization.

The first baseline is \emph{TAP-Net}~\cite{hu2020tap}.  Using the same set of inputs as ours (box states, precedence graph, container height map), it outputs a valid packing sequence of box states. 
The packing position of each box state in the sequence is then computed using a greedy approach, i.e.,  the candidate EMS that yields the highest compactness for the box state is used.  

The second baseline method referred to as \emph{Greedy EMS}, considers all box states and candidate EMS extracted by our objects/container analysis. %, which is one of the well-known heuristic packing strategies~\cite{karabulut2004hybrid}. Given a set of candidate packing positions, the main idea of DBLF is to find the deepest available position (smallest z value) in the layout, at the same time as far as possible to the bottom (smallest y value) and as far as possible left (smallest x value). 
For each $\langle$oriented box, EMS$\rangle$ pair, the packing position is determined using the DBL corners as in our method. 
The pair that yields the highest compactness at each step is greedily selected.
%\rh{Note that when there are multiple pairs leading to the same highest compactness, we will further select the one with the lowest packing height. %\mlc{Double check, as I thought the same compactness will always lead to the same packing height...} 
%If there are again multiple candidates, we will randomly select one.}
%finds the box state packed by the DBLF strategy that reaches the highest compactness.

To focus the evaluation on packing strategies, both baseline methods use the same source objects/container analysis and update objects/container states after each packing step.
Moreover, for a fair evaluation, the object detection step is bypassed, since TAP-Net~\cite{hu2020tap} requires a full observation of all the objects, \rz{with all the source objects axis-aligned with known sizes and blocking relations.}
\change{We use $N_s = 20$ in the \emph{FIX}  and \emph{RAND} source box setting comparisons, while for the \emph{FIX} source box setting, we further specify $N_f = 5$, which means that the 20 objects are a random combination of 5 pre-determined objects.  Moreover, for the \emph{PPSG} source box setting, the boxes are constructed by splitting the space of one target container, so we set  $N_s = 10$ to avoid getting too small boxes.}

\input{figures/compare_single}

\input{figures/tab_single}

\paragraph{Comparison under \textbf{Single} container setting.}	
\change{Table~\ref{tab:tab_single} compares the performances of our network and the baseline methods when packing different groups of source objects into a \emph{Single} target container. 
%As the goal is to fill the given single target container as fully as possible, we evaluate all the methods with compactness $C$.
We can see that our method gets consistently better performance than those two baseline methods.
The improvement varys from 20\% to 30\% when comparing TAP-Net++ to the \emph{Greedy EMS} basesline.
\emph{TAP-Net} gets the worse results as it uses two-step optimization which selects the object states without simultaneously determining its packing location while both TAP-Net++ and the \emph{Greedy EMS} baseline select the packing strategy in one step after constructing the candidate EMS packing location.
%Compared to TAP-Net++ and the \emph{Greedy EMS} baseline, which both select the packing strategy in one step after constructing the candidate EMS packing location, \emph{TAP-Net} uses two-step optimization which leads to sub-optimal solutions and the worse results. 
%When comparing our method to \emph{Greedy EMS} baseline, we can see that the improvement in \emph{PPSG} settings is much more significant than that in \emph{FIX}  and both \emph{RAND}.
%We think this is because the object sizes are mostly the same in the \emph{FIX} setting, which makes the difference between different strategies quite small.

Figure~\ref{fig:compare_single} shows some visual comparisons of the final packing results of different methods, demonstrating that TAP-Net++ provides more compact packing with more objects in the given container. 
In particular, considering the \emph{PPSG} example illustrated in the last row, TAP-Net++ successfully identifies the optimal packing solution with a perfect compactness score of $C=1$. 
The \emph{Greedy EMS} approach selects suboptimal solutions throughout the process, resulting in narrow empty spaces that cannot accommodate the remaining objects. 
Furthermore, since TAP-Net does not explicitly extract candidate packing locations and mask out invalid pairs as the other two methods do, it failed to select the box state that can be properly packed into the container and terminates with the selection of the highlighted object in transparent.
%It shows that, as the task becomes more challenging (container size increases), all approaches see performance drops.  In addition, while TAP-Net performs better than Greedy EMS under smaller container sizes, it losses its advantage when the container size reaches 100.
%Our method, on the other hand, yields consistently and significantly better results under all container sizes.
%and uses fewer containers. Take the first task as an example, TAP-Net++ and TAP-Net use only one container to pack the boxes, while the final packing height of TAP-Net++ is lower than TAP-Net. TAP-Net++ also provides more compact solutions for the remaining there tasks, even though all methods use two containers.

%\input{figures/tab_eval_prec}

\input{figures/tab_multi_all}
\input{figures/tab_multi_last}

\input{figures/compare_multi}

\paragraph{Comparison under \textbf{Multi} container setting.}
Compared to the previous experiment, packing objects into \emph{Multiple} containers is more challenging because more candidate packing locations need to be considered and longer-term planning needs to be performed to pack all the source objects into the target containers. 

Table~\ref{tab:multi_all} and~\ref{tab:multi_last} compare the performances of our network and the baseline methods under \emph{Multi-All} and \emph{Multi-Last} container settings, respectively. 
They show that our method consistently yields better performance than the two baselines as in the \emph{Single} setting. 
In addition, the packing performances under \emph{Multi-All} are consistently better than those under \emph{Multi-Last} because the former offers more flexible packing options and can make use of remaining free space in the previous containers.
We have also observed that the advantage of our method is more pronounced in the \emph{Multi-Last} setting compared to the \emph{Multi-All} setting. We believe this is because \emph{Multi-Last} poses a more constrained optimization problem, making it more challenging to achieve an optimal packing solution.%, so each decision made has more direct impact on the container number with only the last container can be used to pack.

Figure~\ref{fig:compare_multi} presents visual comparisons of results obtained by different methods for the same set of input objects packed under either the \emph{Multi-All} or \emph{Multi-Last} setting. It is evident that TAP-Net++ consistently achieves more compact packing results compared to the other methods.
When comparing the results between the \emph{Multi-All} and \emph{Multi-Last} settings, particularly for TAP-Net and \emph{Greedy EMS}, we can observe that the first container under the \emph{Multi-All} setting is usually filled with more boxes than the first container under the \emph{Multi-Last} setting in all three source settings (\emph{PPSG}, \emph{RAND}, and \emph{FIX}). This difference arises due to the flexibility of packing all available containers in the \emph{Multi-All} setting.}

\input{figures/tab_diff_num}

\subsection{Generality of TAP-Net++}

To evaluate the generality of TAP-Net++, we further test it under different object numbers and container sizes. %, and input workspaces. % and measurement precision. 
Without loss of generality, we here focus the application of packing \emph{RAND} source objects into multiple containers under the \emph{Multi-Last} setting.
%All tests below are repeated on 1,000 randomly generated initial box configurations.

\paragraph{Generality on object number}
The performance of neural networks often degenerates when the training and testing data have different characteristics. Since our approach uses a transformer encoder to encode precedence, it can handle an arbitrary number of input objects at the testing stage. Here we want to find out the impact of using different object numbers under training and testing stages.
\change{When testing the task of packing $N_s = \{20,40,60, 80,100\}$ objects, two versions of TAP-Net++ are compared. The first version is trained using $20$ objects, whereas the second is training using the same number of $N_s$ objects.
As shown in Table~\ref{tab:diff_num}, the object number used for training TAP-Net++ has little impact on its performance. \rh{We also test the inference runtime performance with respect to the number of boxes. With an NVIDIA TITAN Xp, it takes consistently 20ms for TAP-Net++ to infer a <oriented box, EMS> pair with different numbers of boxes.} These suggest the robustness and generality of our method on box number.}

%two versions of TAP-Net++, trained on datasets with $10$ and $20$ boxes, respectively, are tested on 1,000 tasks with $10$, $20$, and  $50$ boxes. For TAP-Net, their proposed rolling-based scheme is applied when the test object number is higher than the training object number. The container size is kept small at $5$ for a fair comparison with TAP-Net.

%Table~\ref{tab:diff_num} shows the performances of two networks on tasks with different numbers of box ($N_{\text{obj}}$). 
%TAP-Net++ outperforms TAP-Net on all tests. In addition, the best performance is obtained under 20 test objects, even when the network is trained using 10 objects. 
%Our hypothesis is that more objects offer additional room for packing optimization and hence lead to better performance.  
%Increasing object number to 50 only slightly reduces performance, which demonstrates TAP-Net++'s strong generality on more complicated tasks.
% \ml{While TAP-Net++ trained on dataset with a larger object number works better, the one trained on fewer number of boxes generalizes well for more complicated tasks and only yields a small performance decrease.}

%The performances of both TAP-Net and TAP-Net++ when trained on 20 and tested on 10 drop compared to those when tested on 20, I think the main reason is we use the setting with unlimited height, which will naturally give higher compactness for more objects.  

%\input{figures/tab_scalability_num}

\input{figures/tab_container_size}

\paragraph{Generality to container size}
\change{By default, we set the target container size as  $100 \times 100 \times 100$. 
To test the generality of our method to the target container size, we directly apply our network trained on the default target container to two new container sizes: $150 \times 100 \times 100$ and $200 \times 100 \times 100$. 
To make the evaluation metric comparable, the number of source objects are scaled accordingly to match target container size changes.
The results are shown in Table~\ref{tab:container_size}, with the performance of \emph{Greedy EMS} baseline included for comparison.
We can see that the performance of our method is very stable, whereas that of the \emph{Greedy EMS} baseline degenerates noticeably. }

\input{figures/real_dataset}

\input{figures/compare_practical}

\subsection{Results on causally stacked boxes}
%Next, we compare the three approaches under practical settings that TAP-Net++ is designed for. 
\change{Thanks to our informative object information encoder and the generality of our network to the object number, the network we trained on axis-aligned boxes with full observation can be directly used to deal with more practical settings with causally stacked boxes without the need for fine-tuning either in the simulator or in the real world. }

\paragraph{Simulated environment}
As shown in Figure~\ref{fig:overview}(a \& d), for each task, we generate $N_{s}$ objects with random poses and let them fall onto a flat surface by gravity through pybullet simulator~\cite{coumans2016pybullet}.
This provides the initial configuration for boxes, which are unknown to TAP-Net++. 
RGBD images are captured using a virtual camera to train Mask RCNN network~\cite{he2017mask} for object extraction.
Once the box shapes and their precedence are extracted, we apply TAP-Net++ trained on axis-aligned boxes with full observation to directly output the packing strategy.
\change{As TAP-Net cannot handle dynamic inputs, we only compare to the \emph{Greedy EMS} baseline; see Table~\ref{tab:tab_generality_practical}. 
We can see that our method performs much better than \emph{Greedy EMS} and the results are similar to the results tested on the axis-aligned boxes with full observation.

Figure~\ref{fig:compare_practical} shows visual comparisons of the final packing results of the two methods under this practical setting. 
For the input boxes casually stacked on the workspace as shown on the left, we found that the object detector can extract correct box sizes and locations for the following packing strategy optimization. 
Compared to the \emph{Greedy EMS} baseline on the final packing results, our method successfully packs boxes into fewer containers with higher compactness. }

\input{figures/tab_generality_practical}

\paragraph{Real world tests}
%\subsection{Real world tests}
Besides experiments in the simulated environment, we also validate our method using a real robot. 
The Mask RCNN network~\cite{he2017mask} is fine-tuned using rendered images of stacked objects with the real workspace setting as the background (see Figure~\ref{fig:real_dataset}), whereas the TAP-Net++ trained using synthesized axis-aligned boxes is again applied directly in real tests without fine-tuning.
Figure~\ref{fig:teaser} shows several keyframes from one packing process. 

\change{Note that our approach may pick up objects from different sides, but always pack them into the container from the top. 
%Hence, the gap required to pack real objects is quite small. 
During the real execution, we enlarge the detected box sizes by a small amount before passing them to the TAP-Net++ to accommodate various errors of executing a packing plan in the real world,}
\rh{which may result in relatively loose packing compared to the simulated results. }
%Figure~\ref{fig:teaser} shows several keyframes from one packing process. 
%Note how randomly stacked objects are tightly packed into the target container. 
%Please refer to the complementary video for more packing examples by the real robot.}

%Please refer to the video for more packing examples by the real robot.
\rh{We found that there are discrepancies in packing performance between the simulation and the real world. 
TAP-Net++ faces distinct complexities when executed in real-world scenarios compared to simulations mainly due to the inaccurate visual perception.
The point clouds acquired by depth cameras in real-world scenarios often contain noises and distortions, and the Mask RCNN trained on rendered images may yield errors when applied to real images due to discrepancies between the virtual and actual environment. 
These factors result in potential inaccuracies in box size estimation and more frequently trigger the algorithm to re-plan placement than in simulation.
Consequently, this might guide the packing strategy towards a sub-optimal solution instead of a globally optimal arrangement.
%Our training process does not have a design to deal with this aspect. 
Incorporating such stochastic placement events into our training process could potentially enhance the robustness and generalization of the algorithm.}

\subsection{Ablation studies on network design}
We now perform ablation studies on several key design choices.
The application of packing \emph{RAND} source objects into multiple containers under the \emph{Multi-Last} setting is again used here.

\paragraph{Precedence encoder}
	In our workspace setting, the number of boxes detected varies at each step, and so is the size of the precedence matrix. 
	We design a precedence encoder based on the transformer to handle this dynamic input size. 
	Another choice would be the RNN, as it can handle dynamic input size as well. 
	Since TAP-Net uses a CNN layer to encode the precedence matrix, it is also tested in this ablation study, even though its input size needs to be fixed. 
	
	The comparison of these three types of precedence encoders is shown in Table~\ref{tab:ablation_precedence} . 
	It shows that our transformer-based encoder has the best performance. 
	Although the performance of CNN is close when the input size is fixed, the transformer naturally supports dynamic input size and is more suitable in our setting.

\paragraph{Packing locations}
\change{Other than the original EMS, we added extra constrained EMS locations to provide more candidate packing locations for stable packing, as shown in Figure~\ref{fig:ems}.
Table~\ref{tab:ablation_location} shows the quantitative comparisons. We can see that with extra constrained EMS locations %\mlc{I felt that the explanation on why more locations are use is unclear.} 
effective candidate packing locations, better solutions can be found to increase performance.
When checking the whole packing process, we indeed observe that compared to using the original EMS locations only, 
TAP-Net++ learns to select the constrained EMS  where the boxes underneath can provide more stable support as long as it provides enough space. }

\paragraph{Packing corners}
	Given a $\langle$oriented box, EMS$\rangle$ pair, TAP-Net++ always computes the packing location by aligning the DBL corners of both. It's possible to try aligning all four bottom corners and pick the best.
	Table~\ref{tab:ablation_corner} shows the performance comparison of using either one or four corners for packing location calculation.
	It shows that the performance of using the fixed DBL corner is slightly better than trying all four corners.
	Our hypothesis is that using a simple and consistent rule for packing location selection helps the network to focus on the optimization for $\langle$oriented box, EMS$\rangle$ pair. This hypothesis is supported by the observation that the network trained using four candidate corners eventually settles down with using the same corner for all EMS.
	
\input{figures/tab_ablation_precedence}	
\input{figures/tab_ablation_location}	
\input{figures/tab_ablation_corner}	
\input{figures/tab_ablation_feasibility}

\paragraph{Feasibility mask}
\change{When taking stability into consideration, we found that only using the matching score to compute the relative scores among different $\langle$oriented box, EMS$\rangle$ pairs is not enough to capture the stability of each individual pair well. Thus, we further compute a feasibility mask using the $\mathbf{sigmoid}$ operator after fusing the features to make the network learn the pair-wise properties better, and then combine the matching score and feasibility mask together to get the final output. 
We found that with the extra branch to predict the feasibility mask, TAP-Net learns to estimate the stability of each packing pair implicitly based on whether this selection increases the number of target containers and thus decreases the average compactness. }
%\rh{We added an extra feasibility mask to estimate the stability \mlc{I don't think how to feasibility mask to improve stability is discussed above.  Simply explain how feasibility mask improves the performance in general is better, I think.  } of each packing pair implicitly based on whether this selection increases the number of target containers. 
\change{Other than adding the feasibility mask to learn the information implicitly, we also tried guiding the network to learn stable packing by adding an \rz{instability} penalty during training. Specifically, once an object is unstable, we add a new container and then add a penalty of $-0.1$ to our reward value.
As shown in Table~\ref{tab:ablation_feasibility}, compared to either directly encoding this information into the matching score or adding a direct penalty for unstable packing without introducing the extra feasibility mask, %\mlc{how do you directly adding penalty?}
our current strategy obtained the best results over all evaluation metrics.}

%\input{figures/tab_ablation_attention}

%\paragraph{Attention module}
%	As shown in Figure~\ref{fig:network}, we use a multi-head attention module in both the source and target encoders to extract core information in the feature map. 
%	The Transformer encoder~\cite{vaswani2017attention} can also be applied here. Table~\ref{tab:ablation_attn} compares the performance of using a multi-head attention module vs the transformer-based encoder. The network with attention module has fewer memory requirements and slightly better performance.

\subsection{Generality on input without precedence} 

\rz{Our paper focuses more on the most challenging TAP problem setting with casually stacked and partially observed objects, where the precedence between the objects has to be taken into consideration.
On the other hand, our method is sufficiently flexible to be directly used to deal with other application scenarios~\cite{zhao2021learning}, where objects are lying on the conveyor belt for packing so that there is no precedence between objects. 
In this case, there may be $m$ objects on the conveyor at the same time, each of which can only be picked up from the top with rotation only allowed around the z-axis, and the goal is to pack as many objects as possible into a given container. 
This is essentially the same as our \emph{RAND} + \emph{Single} setting, where each time, $m$ objects are considered without precedence.} %as input.

\rh{
To show the generality of TAP-Net++ without object precedence, we present a comparison to the SOTA method PCT~\cite{zhao2021learning} under their experiment setting as explained above, where $m$ is set to be 4, and the results are presented in Table~\ref{tab:comp_pct}.
We can see that even though TAP-Net++ was not designed for this setting, it can still achieve slightly better performance than PCT, demonstrating the generality and practicality of our method.}

\input{figures/tab_pct}

%% file: figures/scene_setting.tex
\begin{figure}[!t]
	\centering
	\includegraphics[width=0.98\linewidth]{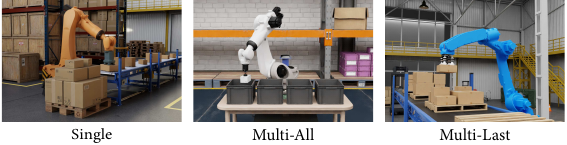}
	\caption{ \change{ %Photographs capturing real-world 
	Packing applications that roughly correspond to different target container settings.} 
%			Photo credits are provided in hyperlinks: \href{https://www.mmh.com/article/robotics_second_wave}{single}, \href{https://www.wired.com/story/amazons-new-robot-sparrow-can-handle-most-items-in-the-everything-store/}{multi-all}, and \href{https://www.mecalux.com/blog/warehouse-robotics}{multi-last}.}
	}
	\label{fig:scene_setting}
\end{figure}

%% file: figures/compare_single.tex
\begin{figure}[!t]
	\centering
	\includegraphics[width=0.98\linewidth]{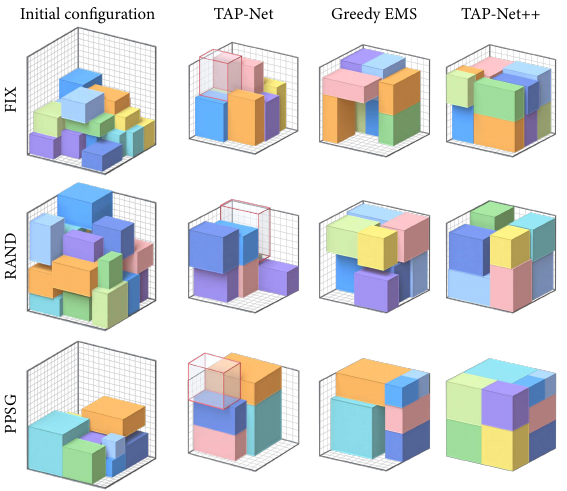}
	\caption{ \change{Comparison of packing results between TAP-Net++ and two baseline methods under \textbf{Single} container setting. 
			For each result obtained using TAP-Net, the box terminates the packing process is shown in transparent with a red frame, which is selected for packing in the current state but cannot fit into the target container using the heuristic packing strategy due to the height limit. Note that this box will not be counted for compactness calculation.} }%\mlc{I thought TAP will still align object by the bottom left corner and drop it, leaving empty space underneath but won't go beyond container boundary...}
	
	%\rh{For each result, we show the box that terminates the packing process in transparent, where the box that cannot fit into the target container is colored with a blue frame and the box that leads to unstable packing is colored with a red frame. 
	%The packing process of remaining results without transparent boxes is terminated due to lack of valid $\langle$oriented box, EMS$\rangle$ pairs. }
	% \jz{The first row shows the packing results under \emph{FIX}., the middle shows the result under \emph{RAND}, the last row shows the results under \emph{PPSG}. }

	%\mlc{I suggest putting the first (fully) packed container on the left side.}
	\label{fig:compare_single}
\end{figure}

%% file: figures/tab_single.tex
\begin{table}[!t]%
	\caption{\change{Comparison on the performance of TAP-Net++ and two baseline methods under the \textbf{Single} container setting, with compactness $C$ as the evaluation metric.}}
	\label{tab:tab_single}
	\begin{minipage}{\columnwidth}
		\begin{center}
			\begin{tabular}{l||l | l | l }
				\hline
				
				 \textbf{Method} &  \textbf{FIX} & \textbf{RAND} & \textbf{PPSG}  \\ \hline 
				 TAP-Net & 0.379 & 0.345 & 0.564 \\ \hline
				Greedy EMS& 0.500 & 0.514 & 0.764  \\ \hline
				\textbf{TAP-Net++} & \textbf{0.657 }& \textbf{0.656} & \textbf{0.929} \\ \hline

			\end{tabular}
		\end{center}
	\end{minipage}
\end{table}%

%% file: figures/tab_multi_all.tex
\begin{table}[!t]%
	\caption{\change{Comparison on the performance of TAP-Net++ and two baseline methods under the \textbf{Multi-All} container setting.}}
	\label{tab:multi_all}
	\begin{minipage}{\columnwidth}
		\begin{center}
			\begin{tabular}{l||l|l|l|l}
				\hline
				\textbf{Input type} &  \textbf{Method} & \bm{$C$} & \bm{$N_t$} & \bm{$\Delta N_t$}  \\ \hline 
				\multirow{3}{*}{FIX} 
				& TAP-Net & 0.441 & 4.24 & 1.95 \\ \cline{2-5}
				& Greedy  EMS& 0.430 & 4.34 & 2.05 \\ \cline{2-5}
				& \textbf{TAP-Net++} & \textbf{0.545} & \textbf{3.42} & \textbf{1.13} \\ \hline
				\multirow{3}{*}{RAND} 
				& TAP-Net & 0.427 & 4.36 & 2.18 \\ \cline{2-5}
				& Greedy  EMS& 0.464 & 4.05 & 1.88 \\ \cline{2-5}
				& \textbf{TAP-Net++} & \textbf{0.538} & \textbf{3.49} & \textbf{1.32} \\ \hline
				\multirow{3}{*}{PPSG} 
				& TAP-Net & 0.478 & 2.13 & 1.13 \\ \cline{2-5}
				& Greedy EMS& 0.560 & 2.02 & 1.02 \\ \cline{2-5}
				& \textbf{TAP-Net++} & \textbf{0.761} & \textbf{1.49} & \textbf{0.48} \\ \hline	
			\end{tabular}
		\end{center}
	\end{minipage}
\end{table}%

%% file: figures/tab_multi_last.tex
\begin{table}[!t]%
	\caption{\change{Comparison on the performance of TAP-Net++ and two baseline methods under the \textbf{Multi-Last} container setting.}}
	\label{tab:multi_last}
	\begin{minipage}{\columnwidth}
		\begin{center}
			\begin{tabular}{l||l|l|l|l}
				\hline
				\textbf{Input type} &  \textbf{Method} & \bm{$C$} & \bm{$N_t$} & \bm{$\Delta N_t$}  \\ \hline 
				\multirow{3}{*}{FIX} 
				& TAP-Net & 0.378 & 5.01 & 2.73 \\ \cline{2-5}
				& Greedy EMS& 0.361 & 5.20 & 2.91 \\ \cline{2-5}
				& \textbf{TAP-Net++} & \textbf{0.535} & \textbf{3.49} & \textbf{1.20} \\ \hline
				\multirow{3}{*}{RAND} 
				& TAP-Net & 0.343 & 5.39 & 3.22 \\ \cline{2-5}
				& Greedy EMS & 0.377 & 5.14 & 2.97 \\ \cline{2-5}
				& \textbf{TAP-Net++} & \textbf{0.540} & \textbf{3.46} & \textbf{1.28} \\ \hline
				\multirow{3}{*}{PPSG} 
				& TAP-Net & 0.455 & 2.27 & 1.27 \\ \cline{2-5}
				& Greedy EMS & 0.544 & 2.14 & 1.14 \\ \cline{2-5}
				& \textbf{TAP-Net++} & \textbf{0.731} & \textbf{1.54} & \textbf{0.54} \\ \hline	
			\end{tabular}
		\end{center}
	\end{minipage}
\end{table}%

%% file: figures/compare_multi.tex
\begin{figure*}[!t]
	\centering
	\includegraphics[width=0.98\textwidth]{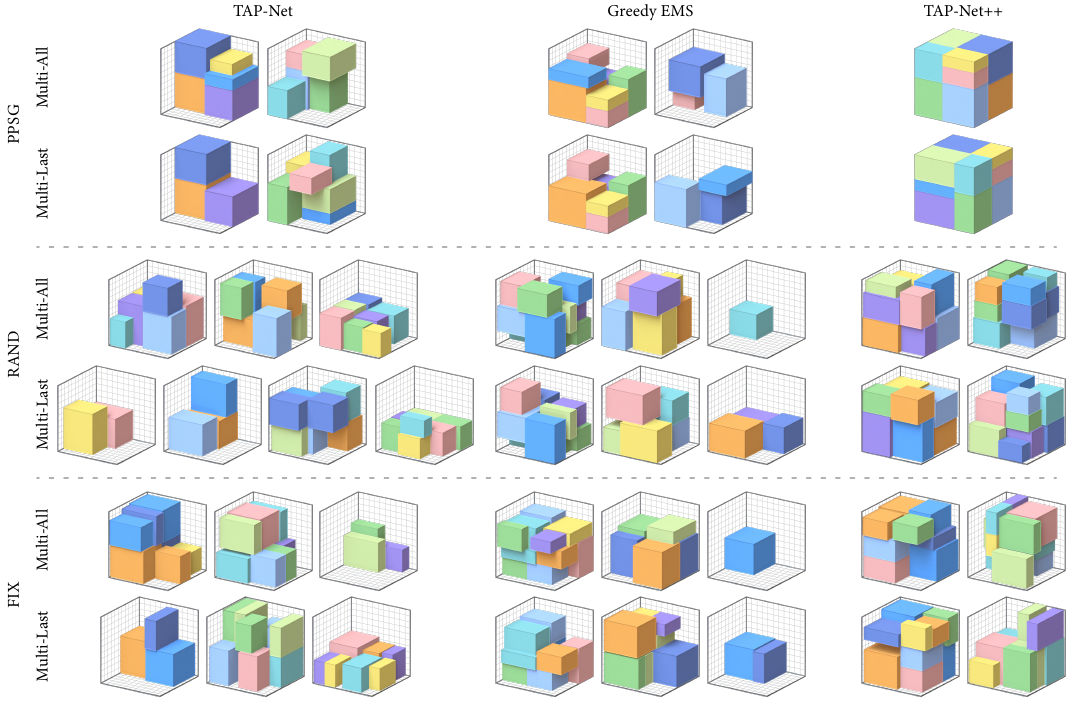}
	\caption{\change{Comparison of packing results between TAP-Net++ and two baseline methods under two \textbf{Multi} container settings.
	For each set of results under the same source setting, the input box configuration is the same but tested under either Multi-all or Multi-Last setting.}
	} 
	%\mlc{I suggest putting the first (fully) packed container on the left side.}
	\label{fig:compare_multi}
\end{figure*}

%% file: figures/tab_diff_num.tex
\begin{table}[!t]%
	\caption{\change{Generality on box numbers $N_s$. When packing $N_s = \{20,40,60, 80, 100\}$ objects, the performance of TAP-Net++ trained on $20$ objects is shown on the left column and that of TAP-Net++ trained on the same $N_s$ objects is shown on the right.}}
	\label{tab:diff_num}
	\begin{minipage}{\columnwidth}
		\begin{center}	
				
		\begin{tabular}{l || l l l  l l l}
			
			\hline
			
			\multirow{2}{*}{\textbf{$N_s$}} & \multicolumn{3}{l|}{Trained on $20$ objects} & \multicolumn{3}{l}{\textbf{Trained on $N_s$ objects}}  \\ \cline{2-7}
			
			&  \multicolumn{1}{l|}{\bm{$C$}} & \multicolumn{1}{l|}{\bm{$N_t$}} & \multicolumn{1}{l|}{\bm{$\Delta N_t$}} & \multicolumn{1}{l|}{\bm{$C$}} & \multicolumn{1}{l|}{\bm{$N_t$}} & \multicolumn{1}{l}{\bm{$\Delta N_t$}}  \\ \hline
			
			20  &  \multicolumn{1}{l|}{0.540} & \multicolumn{1}{l|}{3.46} & \multicolumn{1}{l|}{1.28} &  \multicolumn{1}{l|}{0.540} & \multicolumn{1}{l|}{3.46} & \multicolumn{1}{l}{1.28} \\ \hline
			
			40 &  \multicolumn{1}{l|}{0.556} & \multicolumn{1}{l|}{6.62} & \multicolumn{1}{l|}{2.53}  &  \multicolumn{1}{l|}{\textbf{0.583}} & \multicolumn{1}{l|}{\textbf{6.30}} & \multicolumn{1}{l}{\textbf{2.21}} \\ \hline
			
			60 &  \multicolumn{1}{l|}{0.550} & \multicolumn{1}{l|}{10.03} & \multicolumn{1}{l|}{4.05}  &  \multicolumn{1}{l|}{\textbf{0.586}} & \multicolumn{1}{l|}{\textbf{9.39}} & \multicolumn{1}{l}{\textbf{3.42}} \\ \hline	
			
			80 &  \multicolumn{1}{l|}{0.540} & \multicolumn{1}{l|}{13.59} & \multicolumn{1}{l|}{5.78} &  \multicolumn{1}{l|}{\textbf{0.582}} & \multicolumn{1}{l|}{\textbf{12.60}} & \multicolumn{1}{l}{\textbf{4.79}} \\ \hline	
			
			100 &  \multicolumn{1}{l|}{0.542} & \multicolumn{1}{l|}{16.97} & \multicolumn{1}{l|}{7.33} &  \multicolumn{1}{l|}{\textbf{0.575}} & \multicolumn{1}{l|}{\textbf{15.96}} & \multicolumn{1}{l}{\textbf{6.32}} \\ \hline	
			
		\end{tabular}

		\end{center}
	\end{minipage}
\end{table}%

%\begin{tabular}{l || l l l l  l l  l l }
	
%	\hline
%	\bm{$u$} &  \multicolumn{2}{l|}{20} & \multicolumn{2}{l|}{10} & \multicolumn{2}{l|}{5} & \multicolumn{2}{l}{1}  \\ \hline
	
%	\bm{$N_{obj}$} & \multicolumn{1}{l|}{10} & \multicolumn{1}{l|}{20} & \multicolumn{1}{l|}{10} & \multicolumn{1}{l|}{20} & \multicolumn{1}{l|}{10} & \multicolumn{1}{l|}{20} & \multicolumn{1}{l|}{10} & \multicolumn{1}{l}{20} \\ \hline
	
%	\textbf{TAP-Net} & \multicolumn{1}{l|}{0.769} & \multicolumn{1}{l|}{0.772} & \multicolumn{1}{l|}{0.671} & \multicolumn{1}{l|}{0.681} & \multicolumn{1}{l|}{0.573} & \multicolumn{1}{l|}{0.610} & \multicolumn{1}{l|}{0.557} & \multicolumn{1}{l}{0.526}  \\ \hline
	
%	\textbf{TAP-Net++} & \multicolumn{1}{l|}{\textbf{0.834}} & \multicolumn{1}{l|}{\textbf{0.859}} & \multicolumn{1}{l|}{\textbf{0.744}} & \multicolumn{1}{l|}{\textbf{0.758}} & \multicolumn{1}{l|}{\textbf{0.697}} & \multicolumn{1}{l|}{\textbf{0.709}} & \multicolumn{1}{l|}{\textbf{0.635}} & \multicolumn{1}{l}{\textbf{0.679}} \\ \hline
	
%\end{tabular}

%% file: figures/tab_container_size.tex
\begin{table}[!t]%
	\caption{\change{Generality on container size. The networks (TAP-Net++ and greedy baseline) trained on target container size of $100 \times 100 \times 100$ is tested on different container sizes. TAP-Net++ has very stable performance, whereas greedy baseline uses 60\% more containers when target container size changes.} }
\label{tab:container_size}
\begin{minipage}{\columnwidth}
	\begin{center}	
		
		\begin{tabular}{l || l l l  l l l}
			
			\hline
			
			\multirow{2}{*}{ \textbf{Container size} } & \multicolumn{3}{l|}{\textbf{TAP-Net++}} & \multicolumn{3}{l}{Greedy EMS}  \\ \cline{2-7}
			
			  &  \multicolumn{1}{l|}{\bm{$C$}} & \multicolumn{1}{l|}{\bm{$N_t$}} & \multicolumn{1}{l|}{\bm{$\Delta N_t$}} & \multicolumn{1}{l|}{\bm{$C$}} & \multicolumn{1}{l|}{\bm{$N_t$}} & \multicolumn{1}{l}{\bm{$\Delta N_t$}}  \\ \hline
			
			$100 \times 100 \times 100$   &  \multicolumn{1}{l|}{\textbf{0.540}} & \multicolumn{1}{l|}{\textbf{3.46}} & \multicolumn{1}{l|}{\textbf{1.28}} &  \multicolumn{1}{l|}{0.377} & \multicolumn{1}{l|}{5.14} & \multicolumn{1}{l}{2.97} \\ \hline
			
			$150 \times 100 \times 100$   &  \multicolumn{1}{l|}{\textbf{0.554}} & \multicolumn{1}{l|}{\textbf{3.31}} & \multicolumn{1}{l|}{\textbf{1.20}} &  \multicolumn{1}{l|}{0.286} & \multicolumn{1}{l|}{6.77} & \multicolumn{1}{l}{4.65} \\ \hline
			
			$200 \times 100 \times 100$  &  \multicolumn{1}{l|}{\textbf{0.542}} & \multicolumn{1}{l|}{\textbf{3.42}} & \multicolumn{1}{l|}{\textbf{1.32}}  &  \multicolumn{1}{l|}{0.235} & \multicolumn{1}{l|}{8.34} & \multicolumn{1}{l}{6.25} \\ \hline

		\end{tabular}

	\end{center}
\end{minipage}
\end{table}%

%% file: figures/real_dataset.tex
\begin{figure}[!t]
    \centering
    \includegraphics[width=0.49\textwidth]{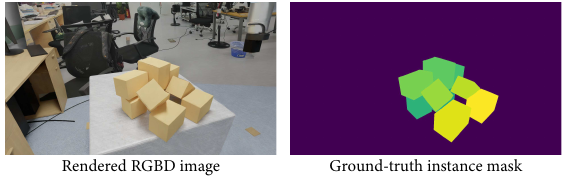}
\caption{ Rendered images used to fine-tune the Mask RCNN network~\cite{he2017mask} for object detection in the real workspace. }
\label{fig:real_dataset}
\end{figure}

%% file: figures/compare_practical.tex
\begin{figure*}[!t]
	\centering
	\includegraphics[width=0.98\textwidth]{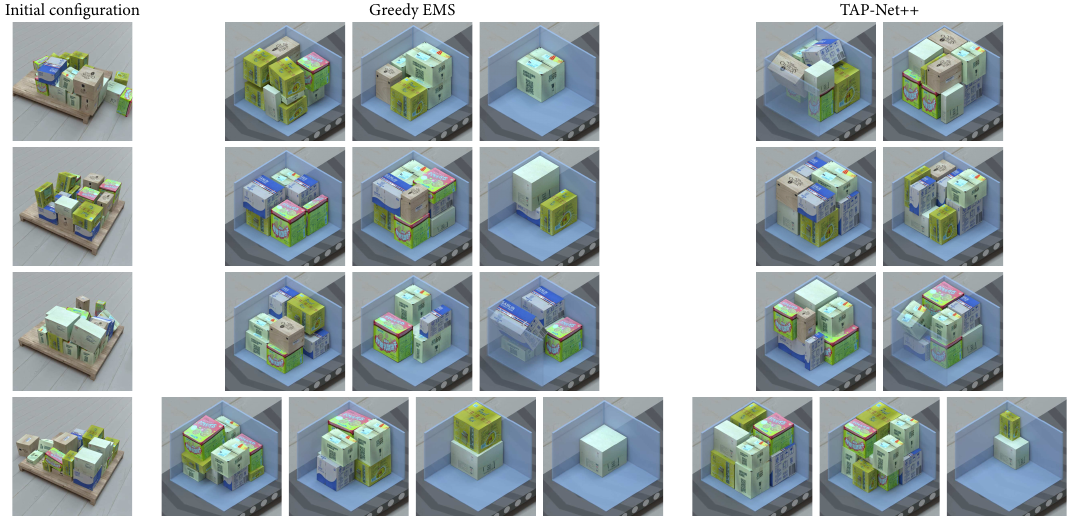}
	\caption{ \change{Comparison of packing results between TAP-Net++ and the \emph{Greedy EMS} baseline in a simulated environment where boxes are initially stacked casually, and their numbers and sizes are obtained through observations.  Note that the container is not fully rendered to show the packed configuration more clearly. } }
	%\mlc{I suggest putting the first (fully) packed container on the left side.}
	\label{fig:compare_practical}
\end{figure*}

%% file: figures/tab_generality_practical.tex
\begin{table}[!t]%
	\caption{\change{Comparison on the performance of the full pipeline (boxes with unknown sizes and numbers are initially stacked casually) obtained using TAP-Net++ and the \emph{Greedy EMS} baseline method under the \textbf{RAND} + \textbf{Multi-Last} setting. }}
	\label{tab:tab_generality_practical}
	\begin{minipage}{\columnwidth}
		\begin{center}
			\begin{tabular}{l||l | l | l }
				\hline
				
				&  \bm{$C$} & \bm{$N_t$} & \bm{$\Delta N_t$}   \\ \hline 
				Greedy EMS &  0.357  &  4.47  &  2.48  \\ \hline
				\textbf{TAP-Net++} & \textbf{0.521}  & \textbf{2.81}  & \textbf{0.86 } \\ \hline
				
			\end{tabular}
		\end{center}
	\end{minipage}
\end{table}%

%% file: figures/tab_ablation_precedence.tex
\begin{table}[!t]%
	\caption{\change{Ablation studies on precedence encoder used for source boxes.}}
	\label{tab:ablation_precedence}
	\begin{minipage}{\columnwidth}
		\begin{center}
			
			\begin{tabular}{l||l|l|l}
				\hline
				\textbf{Precedence encoder} & \bm{$C$} & \bm{$N_t$} & \bm{$\Delta N_t$}  \\ \hline 
				CNN & 0.526 & 3.54 & 1.36 \\ \hline
				RNN & 0.534 & 3.48 & 1.30 \\ \hline
				\textbf{Transformer} & \textbf{0.540} & \textbf{3.46} & \textbf{1.28} \\ \hline
				
			\end{tabular}
			
		\end{center}
	\end{minipage}
\end{table}%

%% file: figures/tab_ablation_location.tex
\begin{table}[!t]%
	\caption{\change{Ablation studies on the set of candidate EMS locations used.}}
	\label{tab:ablation_location}
	\begin{minipage}{\columnwidth}
		\begin{center}
			
			\begin{tabular}{l||l|l|l}
				\hline
				\textbf{Candidate locations} & \bm{$C$} & \bm{$N_t$} & \bm{$\Delta N_t$}  \\ \hline 
				Original EMS & 0.526 & 3.52 & 1.35 \\ \hline
				\textbf{Add constrained EMS} & \textbf{0.540} & \textbf{3.46} & \textbf{1.28} \\ \hline
			\end{tabular}
			
		\end{center}
	\end{minipage}
\end{table}%

%% file: figures/tab_ablation_corner.tex
\begin{table}[!t]%
	\caption{\change{Ablation studies on packing corner selection for box and location alignment.}}
	\label{tab:ablation_corner}
	\begin{minipage}{\columnwidth}
		\begin{center}
			
			\begin{tabular}{l||l|l|l}
				\hline
				 \textbf{Packing corners} & \bm{$C$} & \bm{$N_t$} & \bm{$\Delta N_t$}  \\ \hline 
				All four corners & 0.523 & 3.56 & 1.38 \\ \hline
				\textbf{DBL corner} & \textbf{0.540} & \textbf{3.46} & \textbf{1.28} \\ \hline
					
			\end{tabular}
		
		\end{center}
	\end{minipage}
\end{table}%

%% file: figures/tab_ablation_feasibility.tex
\begin{table}[!t]%
	\caption{\change{Ablation studies on the techniques used for avoiding unstable packing.}}
	\label{tab:ablation_feasibility}
	\begin{minipage}{\columnwidth}
		\begin{center}
			
			\begin{tabular}{l||l|l|l}
				\hline
				\textbf{Method} & \bm{$C$} & \bm{$N_t$} & \bm{$\Delta N_t$}  \\ \hline 
				Matching score only & 0.526 & 3.52 & 1.35 \\ \hline
				+ Unstable penalty & 0.530 & 3.50 & 1.32 \\ \hline
				\textbf{+ Feasibility mask} & \textbf{0.540} & \textbf{3.46} & \textbf{1.28} \\ \hline
			\end{tabular}
			
		\end{center}
	\end{minipage}
\end{table}%

%% file: figures/tab_pct.tex
\begin{table}[!t]%
	\caption{Comparison (compactness $C$) with PCT~\cite{zhao2021learning} under \textbf{RAND} + \textbf{Single} setting, each time considers $4$ objects without precedence. }
	\label{tab:comp_pct}
	\begin{minipage}{\columnwidth}
		\begin{center}
			\begin{tabular}{l||l}
				\hline

				\textbf{Method} & \textbf{RAND}  \\ \hline 
				PCT &  0.663 \\ \hline
				\textbf{TAP-Net++} & \textbf{0.669} \\ \hline
				
			\end{tabular}
		\end{center}
	\end{minipage}
\end{table}%

%% file: 5-conclusion.tex
\section{Conclusion, limitation, and future work}
\label{sec:future}

We develop a learning framework for % \jz{complete}
%real-world 
transport-and-packing using a robotic arm operating in a compact workspace.
As the technical core, TAP-Net++ is trained with reinforcement learning; it takes both source box information and candidate packing locations derived from the target container as input and selects the optimal pair of box state and location to guide the full packing process.
Extensive experiments have been conducted to demonstrate the superiority of our method over the state-of-the-art methods, confirm its generalization capability, and validate several of our key design choices.

One obvious limitation which can be attributed to the visual sensing step is that 
%As a framework for real packing, the input to TAP-Net is extracted from visual sensing, which makes the final packing results depend on the first visual analysis step, for example, 
our method cannot process objects that are undetectable. It may be interesting to explore ways to discover unknown objects using the dynamic visual data captured during the packing process.
Moreover, when applying our method in the real-world test, \change{we currently perform fine-tuning of the object detection network using images captured from a camera positioned above the workspace. While we believe that this is a reasonable requirement considering the fixed nature of the workspace setup, further investigations are needed to explore methods for making the object detection network adaptive to different setups.}

\rz{Furthermore, the inaccuracies arising from object detection and box dimension estimation could lead to more frequent local packing strategy adjustments once an object is picked up and resulting in potentially sub-optimal solutions. Hence, aside from
improving the accuracies, it would also be interesting to incorporate the ensuing uncertainties into the input encoding and enhance the policy learning or add a buffer zone to avoid forced packing.}
%Moreover, our current work does not explicitly account for packing stability, which may cause packed objects to tilt or fall down to invalid the updated container states.
% which will still affect the following packing, although all the workspace states have been updated. 
%A carefully designed stability loss to attain an adequate balance with compactness still need to be developed.
%We did try to add a simple stability score as in TAP-Net but found that the results are still dominated by compactness, so we believe a more accurate stability measure could provide more guidance for the packing.
%\jz{Moreover, we aspire to leverage a physics simulation environment for efficient training, facilitating the algorithm to make more realistic and reliable stability predictions in the real world.}

Last but not least, it is worth exploring ways to optimize low-level control of the robot arm directly through reinforcement learning instead of relying on an existing motion planning method so that the movement paths and packing stability can be simultaneously considered together with compactness.
%candidate EMS are currently extracted from the height map of the target container, 